\documentclass[acmsmall]{acmart}

\AtBeginDocument{%
  \providecommand\BibTeX{{%
    \normalfont B\kern-0.5em{\scshape i\kern-0.25em b}\kern-0.8em\TeX}}}

\setcopyright{acmcopyright}
\copyrightyear{2022}
\acmYear{2022}
\acmDOI{XXXXXXX.XXXXXXX}

\acmConference[CSUR'23]{Make sure to enter the correct conference title from your rights confirmation email}{June 03--05,2023}{Woodstock, NY}
\acmPrice{15.00}
\acmISBN{978-1-4503-XXXX-X/18/06}

\newcommand \ignore[1]{}

\usepackage{url}
\usepackage{hyperref}
\usepackage{colortbl}
\usepackage{multirow}
\usepackage{makecell}

\usepackage{caption}
\usepackage{subcaption}

\begin{document}

\title{Challenges and Opportunities of Few-Shot Learning: A Survey}
\title{A Comprehensive Survey of Challenges and Opportunities of Few-Shot Learning Across Multiple Domains}

\begin{teaserfigure}
  \includegraphics[width=\textwidth]{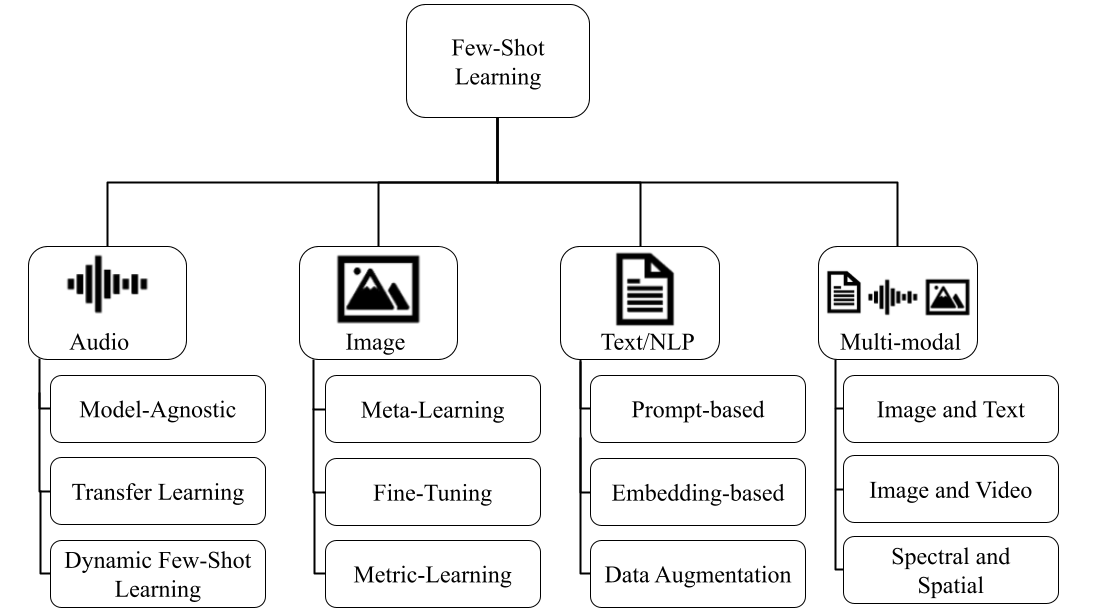}
  \Description{Hierarchy Diagram of Few-Shot Learning}
  \caption{Taxonomy of Few-Shot Learning}
  \label{fig:teaser}
\end{teaserfigure}

\author{Andrea Gajic}
\email{agajic@purdue.edu}
\author{Sudip Vhaduri}
\email{svhaduri@purdue.edu}
\affiliation{%
  \institution{Purdue University}
  \streetaddress{610 Purdue Mall}
  \city{West Lafayette}
  \country{United States}
}

\renewcommand{\shortauthors}{Andrea Gajic, et al.}

\begin{abstract}
In a world where new domains are constantly discovered and machine learning (ML) is applied to automate new tasks every day, challenges arise with the number of samples available to train ML models. While the traditional ML training relies heavily on data volume, finding a large dataset with a lot of usable samples is not always easy, and often the process takes time. For instance, when a new human transmissible disease such as COVID-19 breaks out and there is an immediate surge for rapid diagnosis, followed by rapid isolation of infected individuals from healthy ones to contain the spread, there is an immediate need to create tools/automation using machine learning models. At the early stage of an outbreak, it is not only difficult to obtain a lot of samples, but also difficult to understand the details about the disease, to process the data needed to train a traditional ML model. 
A solution for this can be a few-shot learning approach. This paper presents challenges and opportunities of few-shot approaches that vary across major domains, i.e., audio, image, text, and their combinations, with their strengths and weaknesses. This detailed understanding can help to adopt appropriate approaches applicable to different domains and applications. 
\end{abstract}

\ccsdesc[500]{Computing methodologies~Machine learning approaches}
\ccsdesc[300]{Computing methodologies~Natural language processing}

\keywords{Few-Shot Learning, Audio, Image, Text, Natural Language Processing, Multi-Modal}

\maketitle




\section{Introduction}\label{Intro}
In situations where only small datasets are available initially and decisions have to be made before larger datasets are available, few-shot learning is a critical tool. In few-shot learning, a machine learning model is designed to train with a minimal dataset and then to make reasonably accurate predictions. In situations such as the COVID-19 pandemic and in times of financial crisis, the application of few-shot learning can be beneficial.

In a state of emergency due to a sudden outbreak, e.g., the COVID-19 pandemic, few-shot learning can be used to predict infected individuals quickly and better control the disease spread through human transmissions. That can also reduce the large death toll and associated economic loss. For instance, the 2019 COVID-19 pandemic, caused by the SARS-CoV-2 virus, led to a global healthcare crisis as well as economic disruption and societal crisis due to illness, lockdowns, and vaccine rollouts. By July 2020, an estimated 120,000 people had been affected, and more than 4,000 died~\cite{akbulaev2020economic} due to the limited availability of data, which is needed to develop traditional ML models. With few-shot learning and early development of disease symptom detection models from an early stage of disease outbreak, with a few patients' samples, the situation could be improved.

Few-shot learning can also play a role in an economic recession. When a recession approaches, there are often signs in markets such as the job market and housing market. However, the signs are slow at first, resulting in only a small amount of data available to evaluate if a recession will occur. After the economic downturn in 2007, research was conducted on how it impacted the elderly. It was noted that an estimated 42\% of elderly people were struggling to buy essential items, resulting in reduced or skipped meals~\cite{fenge2012impact}. This had a long-lasting impact on their health, which could possibly
have been prevented. With few-shot learning applied to predict the economic downturn ahead
of time, it could have been possible for people to prepare better, and for governments to support
their people through stimulus packages, subsidies, and investing in job creation. This preparedness may have been able to reduce the struggles with the purchase of essentials such as food and other supplies that many people had.

%
Similarly, few-shot learning can contribute to a variety of areas such as understanding risks~\cite{vhaduri2022predicting,vhaduri2023bag,vhaduri2024mwiotauth} and authenticating users through physiological data~\cite{vhaduri2023implicit,vhaduri2017towards,vhaduri2018biometric,dibbo2021onphone,cheung2020continuous,cheung2020context,muratyan2021opportunistic,vhaduri2021HIAuth}, tracking well-being~\cite{vhaduri2022understanding,vhaduri2015design,vhaduri2017design}, evaluating sleep quality~\cite{chen2020estimating,gomez2024assessing}, monitoring physical health~\cite{vhaduri2016assessing,vhaduri2020adherence} and respiratory disease symptoms~\cite{dibbo2021effect,vhaduri2023transfer,vhaduri2020nocturnal}, managing stress and improving mental health~\cite{vhaduri2021deriving, dibbo2021visualizing,vhaduri2021predicting,kim2020understanding}, and discovering patterns related to places of interest~\cite{vhaduri2018opportunisticTBD,vhaduri2018opportunisticICHI}. 
While this paper presents comprehensive details about few-shot learning approaches applicable to the image, audio, and natural language processing (NLP) domains and their combinations, along with the time and resource requirements, and the opportunities and current and future applications, this paper will guide future researchers and decision-makers to quickly develop domain varying machine learning models from relatively smaller datasets and perform the automation when it is critical to find large data. 
%
%

\noindent{\bf Organization:} 
The paper presents few-shot learning approaches with their shortcomings and opportunities/applications across the audio domain (Section \ref{audio}), the image domain (Section \ref{image}), the NLP/text domain (Section \ref{Text}), and the multi-modal domain (Section \ref{multi}). 
Under each domain, different approaches will be discussed to provide a fundamental understanding of how few-shot learning can be incorporated into the domain, along with time and resource requirements, and current and possible future applications. 
Figure ~\ref{fig:teaser} depicts a taxonomy of the different approaches that are suitable for major domains. 

\section{Preliminaries}\label{prelim}
Before we describe the different few-shot learning approaches, we first define some terminology used in this manuscript.

\noindent{\bf {\em Accuracy} (ACC)} is a performance measure used for assessing what fraction of predictions are correct and it is calculated as: 
\begin{equation}
\label{acc}
ACC = \frac{TP+TN}{TP+FN+FP+TN} = \frac{TP + TN}{P + N}
\end{equation}

\noindent{\bf {\em $F_1$ - Score} ($F_1$)} is a performance measure used for a balanced assessment of a model's effectiveness across both positive and negative instances. 
It combines both precision (PRE) and recall (REC)~\cite{vhaduri2023environment} into a cohesive measure, as shown below.
\begin{equation}
\label{F1}
F1 = \frac{PRE \times REC}{PRE + REC}  \times 2
\end{equation}

\noindent{\bf {\em Dice Similarity Coefficient} (DSC)} is a performance measure used for evaluating the similarity between two samples with a prediction mask {\em m} against the ground truth mask {\em g}. The equation used to compute the DSC is shown below:

\begin{equation}
\label{DSC}
\text{DSC} = \frac{2 |m \cap g|}{|m| + |g|}
\end{equation}


\noindent {\bf {\em Mean Opinion Score ($MOS$)}} is a calculated average of human evaluations where scores are given on a five-point scale by at least five judges.
\begin{equation}
\label{MOS}
\end{equation}

Terms presented in the above equations have consistent meanings in machine learning.
An ideal model should have higher positive performance measures (i.e., Equations~\ref{acc} --~\ref{MOS}).

\section{Audio Domain}\label{audio}
The section below will present a detailed review of the challenges and opportunities of few-shot learning approaches in the audio domain. 

\subsection{Introduction to Few Shot Learning Approaches and Audio}\label{audio_Intro} 
Few-shot learning in the audio domain takes on different forms with different methods. In some cases, few-shot sound detection is applied to classify unseen sounds with only a few support labeled samples to fine tune the model \cite{zhao2022adaptive, wang2020few, wolters2021proposal, chou2019learning}. Another problem that few-shot learning can be applied to in the audio domain is multilingual speech emotion recognition, as there are sometimes not large datasets for languages which are not as widely used as others \cite{naman2022fixed}. Few-shot learning in the audio domain is also developed for the extraction of a target sound from a group of other sounds \cite{delcroix2021few}, such as picking out the violin sound from a concert recording. One noted issue with the audio domain is that labels of a higher quality are more labor-intensive to create \cite{liang2023adapting}, making model development more challenging. There are many different approaches which are taken, and different ways by which they are categorized. Algorithms, including the model-agnostic and meta-curvature, can be classified as overall meta-learning methods that learn tasks to be able to process new and similar ones \cite{zhou2023metarl}. Another few-shot learning algorithm that can be applied is the transfer learning algorithm, which are based on learning from previous models and adapting that knowledge to a new, smaller dataset \cite{pons2019training}. Approaches including the optimization-based model-agnostic algorithms, transfer learning algorithms and dynamic few-shot learning algorithms are taken to fulfill these applications. The aforementioned algorithms, including their successes and possible issues as well as existing models which apply them, will be discussed in the next sections.

\subsection{Few Shot Learning Approaches for Audio}\label{audio_Approach}
The sections below will discuss the various few-shot learning approaches for the audio domain. In Section~\ref{audio_a1}, Model-Agnostic Learning will be discussed. In Section~\ref{audio_a2}, Transfer Learning will be reviewed and and Dynamic Few-Shot Learning will be covered in Section~\ref{audio_a3}. After the approaches are discussed, applications of few-shot learning in the audio domain will be covered.

\subsubsection{Model-Agnostic Meta-Learning}\label{audio_a1}
\begin{figure}[hbt]
\includegraphics[width=4.5in]{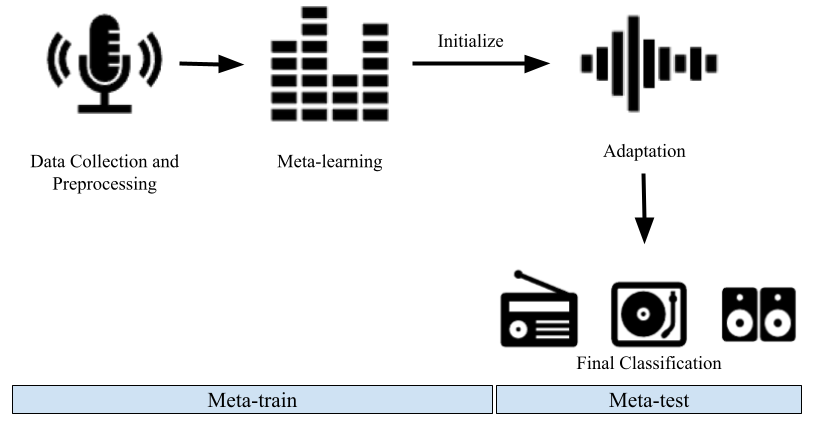}
\caption{Diagram of Model-Agnostic Meta-learning Application for Audio}
\label{fig:MLAudio}
\end{figure}

Meta-learning is an approach that was popularized for image-related tasks and which is being expanded into the sphere of audio tasks. These tasks include audio event recognition, text-to-speech, speaker recognition, speech recognition, and other related areas to speech processing \cite{zhou2021meta}. Meta-learning can design models that adapt to new environments and learn new skills quickly with only a few training samples, making it suitable for solving few-shot tasks \cite{huang2022meta}. The model accomplishes this by using multiple subtasks to learn the parameter initialization so that fine-tuning can be applied to the initialization with only a few labels and still perform well with the targeted tasks \cite{li2021meta}. In each approach, there is a task-independent encoding function and a task-specific classifier in task-independent learning \cite{mittal2021representation}.  ProtoNet is a model for learning embeddings for classification, while Ridge is a model for preventing overfitting, uses linear regression, and MetaOptNet is a model for optimizing feature representations. The models use nearest neighbor, linear regression, and linear SVM, respectively, as classifiers. For SVM-based methods, approximate gradients are captured using implicit functions to facilitate end-to-end training \cite{mittal2021representation}. There are typically two stages in the meta-learning method: the meta-train stage and the meta-test stage \cite{zhang2022adversarial}. During the meta-train stage, the model is trained on base classes using an episodic training strategy \cite{zhang2022task}. In the next stage, the meta-test stage, the network is transferred to previously unseen classes. The transfer is task-agnostic, so the model may not generate the most effective feature representations for distinguishing specific tasks that include novel classes \cite{zhang2022task}. The stages previously described are represented in {Figure~\ref{fig:MLAudio}}. There are also two groups of parameters, meta-parameters, which are randomly initialized, and adapt-parameters, which are adjusted during training based on the learning done \cite{hu2023meta}.

\begin{equation*}
\mathop {\min }\limits_{(x_0, y_0)} \mathcal{L}( (a_N, b_N); \mathcal{R}) = \mathop {\min }\limits_{x_0, y_0} \mathcal{L}( \text{Opt}(x_0, y_0; \mathcal{L}, \mathcal{T}, M); \mathcal{R}) \tag{1}
\end{equation*}

One of the most effective meta-learning methods that has been observed is the model-agnostic meta-learning (MAML) model, which is gradient-based \cite{moon2024task, finn2017model}. In the equation $x_0$ and $y_0$ represent the initial parameters that MAML optimizes, $\mathcal{T}$ is used for task-specific adaptation via gradient descent, ${M}$ is the number of gradient steps taken to update the model, $(a_N, b_N)$ is the adapted parameters after $M$ optimization steps, and $\mathcal{R}$ is the final evaluation set where the loss is minimized. 
Equation 1 above represents the optimization problem for the model-agnostic meta-learning framework for few-shot learning. It is designed to compute based on initialization parameters from a meta-training set so that the model can perform on a query set \cite{shi2020few}. The initialized parameters are updated using M steps of gradient descent based on a loss function and computed over a support set. Notably, the goal of MAML is to minimize cross-entropy loss in the updated model over the query set.

The MAML algorithm learns the initialization parameters from the meta training set so that the model can quickly adapt to new tasks after only a few steps of gradient descent \cite{shi2020few}. The MAML algorithm was applied to a multi-speaker TTS model, and the results of the study conducted on the Meta-TTS model demonstrated that it could generate speech with high speaker similarity with only a small number of samples. As a result, fewer adaptation steps were required than in the speaker adaptation baseline. It also exceeded the baseline level of speaker encoding under the same training conditions \cite{huang2022meta}. The two baselines did not use the MAML algorithm while Meta-TTS did, and the results support that the model using meta-learning was more successful than those that did not. The ground truth was calculated to have a MOS of 4.54 $\pm$ 0.09. After 10 steps, the baseline had an MOS of 1.32 $\pm$ 0.13, while Meta-TTS resulted in an MOS of 3.45 $\pm$ 0.14. The Meta-TTS generates audio similar to the target with far fewer steps than the baseline. Gradient-based meta-learning is a subset of models that can rapidly adapt to new tasks using gradient-based updates, usually employing a bilevel optimization procedure \cite{chen2022learning}. Research has shown that the aforementioned approaches consistently outperform both metric-based and baseline methods \cite{heggan2022metaaudio}. Meta-curvature is a gradient-based method that expands on MAML by learning a transform of the inner optimization gradients to get a better generalization on new tasks.

However, there are drawbacks to MAML. If the source task knowledge is too diverse, it has been noted to fail to generalize the common knowledge of the source tasks \cite{moon2024task}. In the DeepASC model, the team overcame this by accommodating task-wide general knowledge and task-specific knowledge into the MAML-initialized parameters. As such, the DeepASC model was able to become more flexible than the original MAML while still keeping to its core principles.

\subsubsection{Transfer Learning}\label{audio_a2}
\begin{figure}[hbt]
\includegraphics[width=4in]{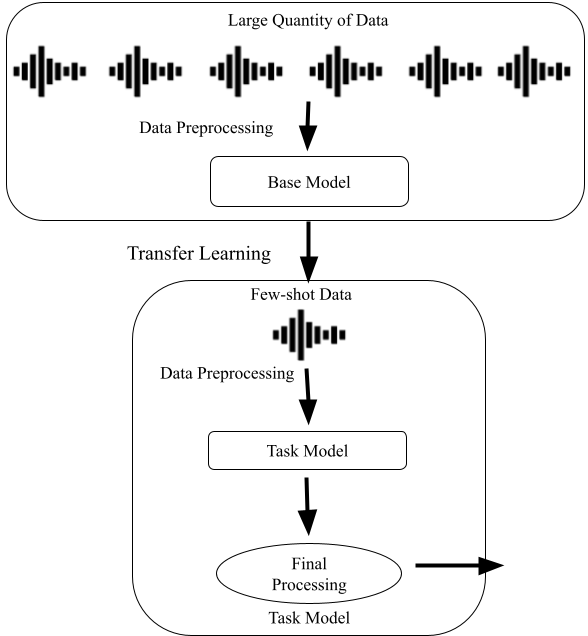}
\caption{Diagram of Transfer Learning Application for Audio}
\label{fig:TLAudio}
\end{figure}

Another approach that can be applied to audio-based few-shot learning is transfer learning. Transfer learning takes advantage of the knowledge learned from other data or features \cite{deng2013sparse, diment2017transfer} to solve given tasks and to ensure that the model is not completely re-trained \cite{lu2024enhancing}. When using a neural network, transferring pre-trained weights can significantly reduce the number of trainable parameters in the target-task model, thus enabling effective learning with a smaller dataset \cite{choi2017transfer}. Transfer learning consists of two tasks: a source task and a target task \cite{zhang2017transfer}. The tasks can be seen in {Figure~\ref{fig:TLAudio}}. The source task uses existing training data, of which there is an abundance. The target task has a limited amount of data, and any knowledge learned in the source task that may be useful is transferred to the target task \cite{diment2017transfer}.  The transfer-learning approach has been used in the construction of acoustic models for low-resource languages, the adaptation of generative adversarial networks, and the transference of knowledge from the visual to the audio domain \cite{pons2019training}. It has been noted that while with visual-based few-shot learning it is standard to reuse networks which are pre-trained on large datasets; that is not the case with audio-based few-shot learning. Diment and Virtanen suggest that, unlike image data where the early layers of a convolutional neural network can be visualized to reveal basic shapes and textures, applying the same approach to audio data is more challenging \cite{diment2017transfer}.

\begin{equation}
\phi' = \phi - \gamma \nabla_{\phi} \mathcal{L}_{A}(\text{audio}) \left( [ \Psi; \phi ], \Xi_T \{1, 2\} \right)
\tag{2}
\end{equation}

Equation 2 overall represents the process by which the data is processed. The task is trained on its support set in order to carry out the transfer-learning process \cite{liu2024fault}. The equation adjusts the pre-trained model's parameters to better classify new audio classes with a limited amount of data. The $\phi'$ represents the updated parameters of the model after a single gradient step, $\phi$ represents the current parameters of the model pre-update, $\gamma$ is the learning rate which controls the step size during the gradient descent update, and $\nabla_{\phi}$ is the gradient operator. Furthermore, the $\mathcal{L}_{A}$ represents the loss function evaluated on the audio data. This is the difference between the model's predictions and the actual labels on the audio samples. The $[ \Psi; \phi ]$ is a concatenation of two parameter sets, with $\Psi$ as the feature extractor and $\phi$ as the classifier parameters, which take the extracted features and output the final predictions. The last section, $\Xi_T \{1, 2\}$ is the support set for task $T$, which contains audio samples from class 1 and class 2. 

 It has been noted that a basic transfer learning approach becomes increasingly effective when the number of support samples grows \cite{wang2022active}. Wang et al. found that retraining a supervised model with additional novel examples led to the highest mean average predictions (mAP) on both base and novel classes, while also achieving an $F$-measure comparable to other methods \cite{wang2021few}, likely due to the use of transfer-learning. A novel method utilizing hierarchical long- and short-term memory neural networks has been proposed to solve an emerging overfitting problem \cite{lu2024enhancing}. This approach integrates the transfer-learning framework and accelerates the converge speed of the task model. It also helps maintain the sensitivity to input data while minimizing overfitting when working with a limited number of new task samples \cite{lu2024enhancing}. Another proposed solution for overfitting is using transfer learning with deep networks. One deep convolutional neural network in the image domain, called AlexNet \cite{krizhevsky2017imagenet}, has been applied with transfer learning to audio tasks.  In the case of AlexNet, the proposed model is a CNN (convolutional neural network) model which consists of a pre-trained AlexNet, where the preserved parameters of the pre-trained model serve as initialization for training \cite{lu2021detection}. The trained classification model was able to accurately identify different whale sounds with a low test time of 9.5 seconds and with $F_1$ scores varying from 0.9533 to 0.9850. It should be noted however, that CNN-based methods are costly and are sensitive to hyperparameters. 

\subsubsection{Dynamic Few Shot Learning}\label{audio_a3}
Dynamic few-shot learning (DFSL) was first proposed for dynamic few-shot visual learning by Gidaris and Komodakis for the purpose of few-shot class-incremental learning in the visual domain \cite{gidaris2018dynamic}. Since them, the same approach has been proposed to be applied in the audio domain by Wang et al. for the purposes of few-shot continual learning \cite{xie2023few}. DFSL was optimally designed to solve few-shot problems because it uses an additional episodic training stage in order to train the few-shot weight generator \cite{wang2021calls, xie2023few}. The goal of DFSL is to learn categories while using only a few labeled points and factoring in the base categories with which the model was initially trained. DFSL is implemented to gradually expand the model's capabilities, allowing it to adapt dynamically to new classes while retaining knowledge of previously learned ones~\cite{raimon2024meta}.

\begin{equation}
    w_{z + 1}^* = {\psi _{avg}} \circ {y_{avg}} + {\psi _{att}} \circ w_{att}^*.
\tag{3}
\end{equation}

The novel class weight vector is calculated using Equation 3, where the averaged feature vector is calculated based on the labeled samples from the novel class, and the attention-based weight vector incorporates past knowledge from base classes. The ${y_{avg}}$ represents the averaged feature vector, the ${w_{att}^*.}$ represents the attention-based weight vector that incorporates past knowledge from base classes, and the ${\psi _{avg}}$ and ${\psi _{att}}$ are learnable, while the $\circ$ represents the Hadamard product. 

\begin{equation}
   w_{att}^*.  = \frac{1}{M} \sum\limits_{j = 1}^{M} \sum\limits_{c = 1}^{P} A \, tt\left( {\Theta_r}{y_j}, {m_c} \right) \cdot v_c,
\tag{4}
\end{equation}

The equation above includes $\frac{1}{M} \sum\limits_{j = 1}^{M}$ which iterates over ${M}$ novel samples and computes an average contribution from each novel example to the weight vector. The $\sum\limits_{c = 1}^{P}$ sum iterates over ${P}$ base classes and results in each novel example's weight vector being influenced by all base class weight vectors. The attention mechanism, $A \, tt\left( {\Theta_r}{y_j}, {m_c} \right)$ determines how much weight should be given to each base class weight vector, $v_c$, when $w_{att}^*.$ is being constructed. The feature vector of a novel class sample is $y_j$ and is transformed using a learnable matrix, $\Theta_r$. The learnable key vector for each class is $m_c$ while cosine similarity measures how well ${\Theta_r}{y_j}$ align with $m_c$. A softmax function normalizes the scores across all base classes so that the end sum is 1. The attention score of each base class determines how much the weight vector, $v_c$ contributes to the novel weight vector.  
Equation 4 shows how to calculate the attention-based weight vector. Cosine similarity is used to compare the transformed feature vector with assigned key vectors, and a softmax function is applied to normalize weights across the base classes \cite{wang2021few}. 

\begin{figure}[hbt]
\includegraphics[width=4.5in]{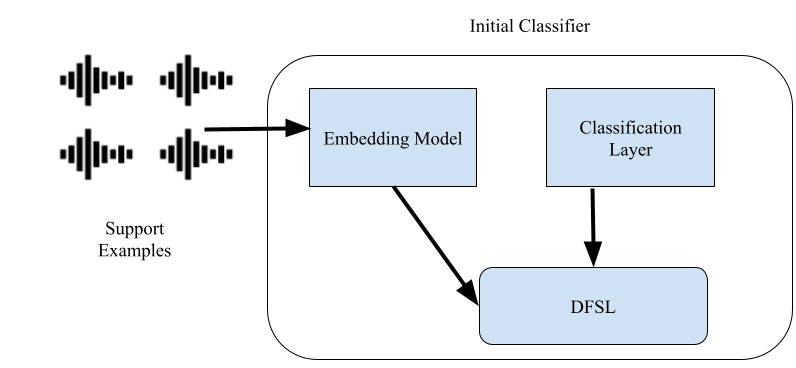}
\caption{Diagram of Dynamic Few-Shot Learning Approach for Audio}
\label{fig:DFSLAudio}
\end{figure}

 A diagram of a high-level overview of DFSL can be seen in {Figure~\ref{fig:DFSLAudio}}. The support examples are sent into an embedding model, and the results are sent to the DFSL processor. The classification layer also provides input to the DFSL processor, which then produces the classification of novel tasks.

One variation of DFSL utilized the prototype of a novel class while leveraging prior knowledge from the classifier by integrating an attention mechanism based on cosine similarity over the base classification matrix \cite{wang2021calls}. Based on research that was conducted to test the proposed methodology in the paper, DFSL performed the best out of all the methods tested in a standard few-shot learning scenario, where the n value was 5 \cite{wang2021calls}. The conclusion drawn from the results was that if minimizing user labeling efforts or computational resources is a priority, DFSL with a smaller $n$ is the preferred approach \cite{wang2021calls}. In the paper, Few-Shot Continual Learning for Audio Classification, it is proposed that a convolutional neural network extracts feature vectors from the audio inputs and then uses classification weight vectors to compute the class likelihoods, which are enhanced by DFSL \cite{wang2021few}. In the proposal, DFSL combined the previous work and a weight generator to create weight vectors for the novel classes using only a few examples. Another methodology of applying DFSL to audio-based few-shot learning is Bioacoustic Event Detection with DFSL (BED-DFSL). The approach is an expansion of DFSL which focuses on detecting specific target events instead of just audio classification \cite{wu2023few}. In each new recording, only one target class is in the interest events category. Furthermore, DFSL is extended into positive and negative binary classifications to detect the specified target events. The BED-DFSL approach has two stages: supervised learning to train a feature encoder using standard cross-entropy loss and a continual learning stage to obtain the parameters of an attention-based weight generator \cite{wu2023few}. The findings indicate that the BED-DFSL and prototypical networks are complementary, with an overall $F$-measure on the validation set reaching 0.606. When compared with a prototypical network run on the same set, BED-DFSL performed with an $F$-measure of 0.543 while the network had an $F$-measure of 0.444. As with other methods, overfitting can occur when standard fine-tuning is applied. One proposed solution is to combine unchanging feature extraction with a query-time adaptive weighting, which combines stability with dynamic adaptation and was noted an area warranting further exploration \cite{nolasco2023learning}. The DFSL approach is a notably new and evolving one in the audio domain, with more work to come. 

\subsection{Time and Resource Considerations}\label{audio_Costs}
Few-shot learning has benefits but poses resource and computational costs. Model-agnostic meta-learning (MAML) exceeds transfer learning in computational expense due to MAML's nested training, which requires costly second-order derivative computations \cite{chauhan2022exploring, tang2022deep, mozafari2022cross, tak2024enhancing, abbas2022sharp} and gradient updates \cite{lin2023model}. These derivatives increase memory usage for storing intermediate gradients \cite{kulkarni2023optimisation} and extend compute time for task simulations. Previous studies have observed that MAML's latency can reach up to 120 seconds for some datasets, making it the most expensive model, while MetaOptNet lags due to linear SVM application \cite{chauhan2022exploring}. First-order approximations reduce computational costs but compromise accuracy and still demand significant resources for diverse task datasets and prolonged training. As a result, MAML is less practical for resource-constrained environments.

Transfer learning is the most efficient for end users \cite{transferCost}. This approach is computationally efficient because it leverages existing feature representations, thus requiring only minimal updates to the model's weights for the new task \cite{tsalera2021comparison}. The resource cost is also relatively low, as it avoids training from scratch and primarily demands a modest amount of labeled data, along with moderate computational power. However, its adaptability is limited by similarity between the pre-training and target tasks, which can reduce effectiveness for divergent domains. 

Dynamic few-shot learning is a less standardized category and encompasses a variety of methods, such as memory-augmented networks and task-specific parameter generation. It falls between MAML and transfer learning in terms of resource demands. The costs vary depending on the approach utilized, but typically, elements of meta-learning and dynamic feature-extraction may be combined. DFSL requires moderate-to-high resources, including memory for storing task representations and computational power for adjustments, while also offering greater flexibility than transfer learning and lower overhead than MAML's meta-training.

Overall, transfer learning is the most resource-efficient due to its reliance on pre-trained models and minimal retraining, while MAML incurs the highest computational cost due to its meta-optimization demands. Dynamic few-shot learning offers a middle ground, with costs varying based on the specific implementation. 

\subsection{Applications}\label{audio_Applications}
\begin{itemize}
    \item In the security realm, few-shot learning can be applied to speech verification in authentication \cite{won2024metric}. A person configuring voice authentication on their device would provide a small number of audio samples. These audio samples would form the small database being used. With few-shot learning, different algorithms could be applied to ensure that, although there is a small database available, the authentication is still accurate. Work has been done on speech verification for short utterances, where the goal was to accept or reject identity claims a speaker made based on only a few enrollment utterances \cite{wang2023utter}.
    \item An example in which few-shot learning can be applied in the audio domain is to classify rare or obscure animal species \cite{moummad2024mixture, moummad2024regularized, nolasco2023learning}. As previously discussed, MAML can be combined with DeepASC to allow it to acquire general knowledge from different ASC source tasks and effectively adjust to the specific characteristics of the target ASC task  \cite{moon2024task}.
    \item Transfer learning can be used in multiple different applications in regards to the audio field. For example, transfer learning could be applied to speech emotion recognition. In this case, the labeled data set could consist of acted speech that has been classified based on previous human efforts \cite{deng2013sparse}. Another application for a transfer-learning approach in the audio domain is that of identifying and classifying different marine mammal sounds \cite{lu2021detection}. This was done with the use of a modified model, AlexNet. The pre-trained part of the model was fine-tuned to the new task and the part which was replaced was replaced with a CNN-based method. 
\end{itemize}

\ignore{-----------------------------------------
\begin{table}[]
\centering
\caption{Summary of different Federated Learning approaches for Image}\label{}
\begin{tabular}{ |c|c|c| } 
\hline
Approach & Algorithms & Performance \\
\hline
\multirow{3}{6em}{Adaptive Federated Learning} & ADAGRAD~\cite{zeiler2012adadelta, duchi2011adaptive} & 38.01 ACC~\cite{zeiler2012adadelta}, 0.044 WER~\cite{duchi2011adaptive} \\ 
& ADADELTA~\cite{zeiler2012adadelta}& 37.03 ACC~\cite{zeiler2012adadelta} \\ 
& CNN~\cite{deng2020adaptive, tam2021adaptive} & 99.94~\cite{tam2021adaptive}  \\ 
\hline
\multirow{3}{6em}{Personalized Federated Learning} & RSENET-34~\cite{gao2021transfer, sinha2019thin}& 99.22 ACC~\cite{gao2021transfer}, 99.48 $F_1$ ~\cite{gao2021transfer} \\  
& MobileNet-v1 & 3.31 WER~\cite{baykara2024fhauc}, 99.93 ACC~\cite{baykara2024fhauc} \\& ReLU~\cite{ioffe2015batch} & 72.20 ACC~\cite{ioffe2015batch} \\

\hline
\multirow{3}{6em}{Collaberative Federated Learning} & CNN~\cite{mabrouk2023ensemble} & 96.63 ACC~\cite{mabrouk2023ensemble}    \\ 
& LEL~\cite{mabrouk2023ensemble, chen2020online} & 96.42 ACC~\cite{mabrouk2023ensemble}, 94.49 PRE~\cite{chen2020online}  \\ 
& GEL~\cite{mabrouk2023ensemble} & 93.55 ACC~\cite{mabrouk2023ensemble}  \\ 
\hline
\end{tabular}
\end{table}
-----------------------------------------}

\section{Image Domain}\label{image}
The section below will present a detailed review of the challenges and opportunities of few-shot learning approaches in the image domain. 
\subsection{Introduction to Few-Shot Learning Approaches and Image}\label{image_Intro}
In the case of the image domain, the goal of the few-shot learning is to accurately identify new categories using only a small number of examples from each category \cite{das2019two}. As such, few-shot learning in the image domain is largely applied for image classification where few-shot learning is applied to classify unseen classes with only a limited number of labeled samples available \cite{chen2021self, chen2019closer}. Many different approaches can be applied to few-shot learning in the image domain. The next sections discuss the overall meta-learning approach, the focused fine-tuning transfer learning approach, and the metric-learning approach, and include a high-level overview, different models that apply them, and benefits as well as drawbacks. The first approach discussed will be the overall meta-learning approach, also known as the learning-to-learn approach, which works to have the model adapt quickly to new tasks in the training stage \cite{jiang2022multi}. The next approach to be discussed is fine-tune transfer learning - a method that utilizes task-specific support images to fine tune the feature extractor network \cite{bateni2020improved}. Metric learning will be the final approach covered, which tries to solve the problems in the image domain by positioning new classes within a metric space, such as Euclidean or cosine distances, in a way that allows for clear class separation \cite{sun2020few}. 

\subsection{Few Shot Learning Approaches for Image}\label{image_Approach}
In the sections below, multiple different few-shot learning approaches in the image domain will be introduced.  Meta-Learning will be covered in Section~\ref{image_a1}, and Fine-Tuning Learning will be covered in Section~\ref{image_a2}. The last approach covered will be Metric-Learning in  Section~\ref{image_a3}. Following the approaches, an application section will cover the various applications of few-shot learning in the image domain. 

\subsubsection{Meta-Learning}\label{image_a1}
\begin{figure}[hbt]
\includegraphics[width=4.5in]{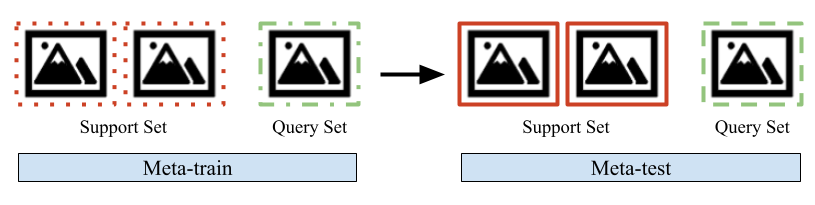}
\caption{Diagram of Meta-Learning Approach for Image Domain}
\label{fig:MLImage}
\end{figure}

In the visual domain, the meta-learning technique is notably prevalent, with the primary goal of developing an algorithm capable of adapting to a variety of tasks \cite{lee2023xdnet}, in either the optimization-based, metric-learning, or model-based methods \cite{kumar2020designing}. Some of the methods under these classifications are designed to be task-agnostic, allowing them to generalize across various domains such as classification, regression, and reinforcement learning, while others are specifically tailored for classification tasks \cite{lee2023xdnet}. These have been developed in the image classification content but for natural images and not industrial-type images with pixel-level defects \cite{lee2023xdnet}. Most existing work has focused on simple image classification tasks under artificial conditions, utilizing small-scale datasets like mini-ImageNet and working with structured problems such as five-way classification \cite{wang2019meta}. Meta-learning models make use of episodic training where the episodes are generated by randomly sampling a subset of the classes and a subset of examples per class \cite{wang2022incremental}. This produces a support and query set to which a classification loss is applied \cite{liu2020few}. This in turn enhances the model’s robustness and optimizes its embedding across different classification tasks. Two common meta-learning methods are the metric-based and optimization-based methods \cite{wang2019hybrid, askari2024enhancing}. Optimization-based methods approach few-shot learning by treating it as a learning-to-learn problem, effectively optimizing model parameters for new tasks \cite{zhang2020few}. The metric-based methods take the learning-to-compare approach, in which the core principle is to train a feature extractor that maps raw input data into a meaningful representation. This transformation ensures that, when represented in the feature space, query and support samples can be easily compared for classification \cite{zhang2020few}. One approach for image-based few-shot learning is Model-Agnostic Meta-Learning (MAML), a learning-to-learn framework inspired by human intelligence \cite{singh2021metamed}. Although MAML-based approaches are standard, another approach that can be applied is the Task-Agnostic Meta-Learning (TAML), which aims to eliminate bias and enhance generalizability by incorporating entropy-based and inequality-minimization concepts \cite{phaphuangwittayakul2021fast}. Meta-learning consists of the meta-train and meta-test stages, most prominently \cite{chi2022metafscil}, and is demonstrated in {Figure~\ref{fig:MLImage}} where the support and query sets are shown to have roles in both stages. 

\begin{equation*}
\min_{\psi, W} \sum_{\tau_j \in \mathcal{T}_B} \mathcal{L}(\mathcal{D}_{\tau_j, \text{test}}; W_j, \psi)
\tag{5}
\end{equation*}

Equation 5 shows a summation of the meta-training model. In the equation, $W$ represents the base learner's parameters, $\psi$ represents the adaptation approach parameters, $\mathcal{T}_B$ represents the mini-batch of sampled tasks, $\mathcal{L}$ represents the meta-loss function, $\mathcal{D}_{\tau_j, \text{test}};$ represents the target dataset for the task $\tau_j$, and $ W_j$ represents the updated parameters after the task-specific adaptation. The training of a meta-learning model involves task sampling, task-specific adaptation, and the meta-training itself \cite{zhu2020multi}. The goal is to optimize the meta-learner so that it adapts to new tasks with limited data. 

The model-agnostic meta-learning algorithm may falter in certain scenarios, such as object detection. This is because the number of parameters in an object detection model is typically greater than that of the model used to assess the effectiveness of MAML \cite{cheng2021meta}. Other works focusing on object detection include Meta-DETR, which discards region-wise prediction and instead works to combine the meta-learning of localization and classification at the image level. It uses a category-agnostic decoder, leveraging global contexts and the synergistic relationship of the two sub-tasks to achieve superior performance \cite{zhang2021meta}. The model addresses the limitation of other object-detection meta-learning frameworks that rely on well-located region proposals, which are challenging to obtain in a few-shot learning environment. Notably for the research, the benchmarks indicate that Meta-DETR significantly outperforms other state-of-the-art methods \cite{zhang2021meta}.

Many different models have been created based on the foundations of the meta-learning approach. A proposed open-set recognition model using meta-learning makes use of episodic training, where a set of new classes is randomly chosen for each episode, and a loss is introduced to maximize the posterior entropy for samples from the classes \cite{liu2020few}. This method better accounts for unseen classes and generates high-entropy posterior distributions that go beyond the regions of the target classes \cite{liu2020few}. The combination of robust meta-learning embeddings and generalizing open-set recognition methodologies allows for performance to improve even in large-scale settings. With deep learning models, it is difficult to adapt them across domains using only limited labeled data. A meta-learning-based solution to this issue is the XDNet framework. It is driven by general feature extractors and meta-learning frameworks to adjust to various domains using classifiers that do not rely on specific parameters, even with limited computational resources \cite{lee2023xdnet}. The MVTec dataset is repurposed to allow for cross-domain generalization efficacy validation and contains data with multiple defect types in an industrial context, which are frequently used as a benchmark for industrial visual inspection tasks \cite{lee2023xdnet}. It has also been noted that existing research has focused on small-scale datasets, and any relevant studies have merely added a few-shot classifier to a detection model without properly addressing few-shot localization, resulting in suboptimal performance \cite{wang2019meta}. A proposed solution is a meta-learning framework that uses meta-level knowledge about model parameter generation from base classes to help create a detector for new classes using only a few examples \cite{wang2019meta}. The MetaDet framework leveraged the Faster R-CNN model, which utilizes the region proposal network and convolutional features as category-agnostic components. The framework can also integrate other detectors such as YOLO, all of which aim to improve the detection performance in few-shot scenarios where few-shot localization is necessary. Another meta-learning model is the SSM-SAM method, which utilizes a rapid online meta-learning optimizer to achieve high performance with limited training data and can be seamlessly integrated into other frameworks \cite{leng2024self}. Compared to other methodologies, the SS-SAM model performed with a DSC result of 0.9309, while other methods tested performed in the range of 0.7748 - 0.9201. In most cases, SS-SAM greatly outperformed the other models. 

Through conducted research it was noted that mini-batch training was advantageous for few-shot learning because it aligns with meta-testing, which is not true in large scale scenarios \cite{liu2020few}. Furthermore, the randomization of the classification tasks, which are learned per episode, forces the embedding f($\phi$) to more effectively generalize to unseen data  \cite{liu2020few}. In few-shot learning, the training data doesn’t have good coverage for variations in classes, so the aforementioned properties make meta-learning a good solution when it comes to few-shot learning. It should however be noted that there are two challenges with current methods. Most existing methods fail to account for time and resource efficiency or budget, which imposes limits on the methods' applicability in different scenarios \cite{chen2021metadelta}. The other challenge is that the success of current methodologies depends heavily on precise hyperparameters \cite{chen2021metadelta}.

\subsubsection{Fine-Tuning Transfer-Learning}\label{image_a2}
\begin{figure}[hbt]
\includegraphics[width=4.5in]{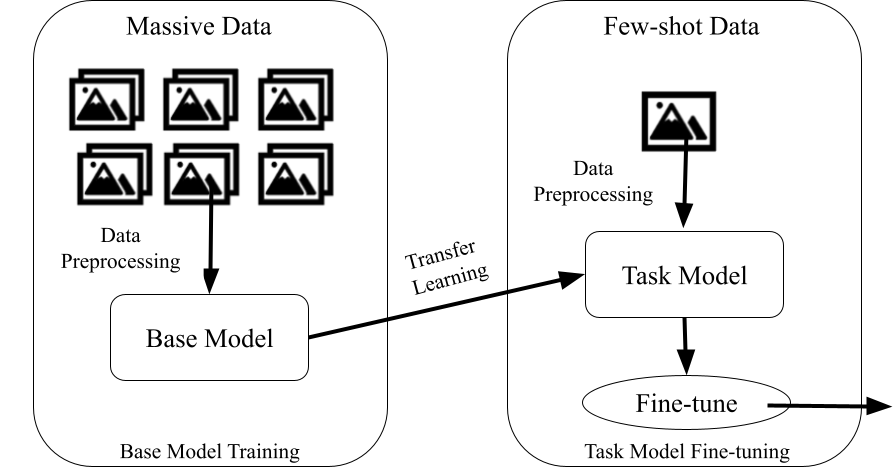}
\caption{Diagram of Fine-Tuning Transfer Learning Approach for Images}
\label{fig:FTLAudio}
\end{figure}

Transfer learning-based few-shot learning is trained on a large dataset and then refined on a smaller, labeled dataset of images. This fine-tuning process enables the model to adapt to the unique characteristics of the new dataset, improving its ability to accurately detect targets in images \cite{uskaner2024efficient, zhang2024few, choi2023incremental}. Transfer learning in the past focused primarily on selecting source domain instances that resemble the target domain instances for transfer, but did not directly enhance the learning performance in the target domain \cite{liu2018few}. The focus of the transfer-learning approach is to focus on transferring knowledge from a secondary domain with the end goal of reducing the amount of labeled data that is required to train a model \cite{rostami2019deep}. An observed baseline that has outperformed state-of-the-art algorithms on multiple benchmarks and different few-shot protocols outperforms all state-of-the-art algorithms on all standard benchmarks and few-shot protocols \cite{dhillon2019baseline} involves pre-training a model on a meta-training dataset using cross-entropy loss, followed by transductively fine-tuning the model on a few-shot dataset. In testing, it was indicated that the accuracy for 1-shot and 5-shot learning with transductive fine-tuning was typically 2\% - 7\% better than the standard models that are considered state of the art. In the f-shot setting, there was typically an increase of 1.5\% - 4\% accuracy compared to other models. Furthermore, a common strategy in deep learning is to fine-tune a pre-trained network model such as ImageNet on a new dataset \cite{guo2019spottune}. A diagram of the fine-tuning approach is shown in {Figure~\ref{fig:FTLAudio}}. Data is processed and passed into the base model, and what is learned there is applied to the task model to assist with the creation of the classification model. The model is fine-tuned and then classification tasks can be carried out.

\begin{equation}
\tilde{c}_j = \frac{1}{\lvert T_j \rvert} \sum_{(a_m, b_m) \in T_j} g_{\phi_{\text{opt}}}(a_m)
\tag{6}
\end{equation}

The $\tilde{c}_j$ is representative of the class prototype vector, which is a reference point for classification in the embedding space. The $\frac{1}{\lvert T_j \rvert}$ is the normalization factor, and denotes the number of support samples belonging to class $j$. The $\sum_{(a_m, b_m) \in T_j}$ represents the summation over the support set where $a_m$ is the input feature and $b_m$ is the label. The last section, $g_{\phi_{\text{opt}}}(a_m)$ represents the feature transformation function. The $g_{\phi_{\text{opt}}}$ maps the input to an embedding space, and $\phi_{\text{opt}}$ represents the optimized parameters of the feature extractor, which has been fine-tuned. 
The equation above, Equation 6, gives an example of the math behind applying fine-tune transfer learning with a prototypical network. The CNN, which is first used, is fine-tuned, and the updated embeddings are utilized for classification. The embeddings are then elevated to create class prototypes, which are computed as the average of the generated embeddings for each class in the defined support set \cite{sarwar2025optimizing}.

A noted problem in transfer learning for few-shot learning is negative transfer. A negative transfer occurs when knowledge transferred from the source domain to the target domain adversely affects performance \cite{liu2018few}. A proposed model that addresses these issues is the analogical transfer learning model, which assumes the target and source domains are related, but not identical. The revision stage of the model identifies and uses source instances that contribute to the target hypothesis and eliminate inconsistent knowledge. The transfer stage then applies the refined source hypothesis alongside the target hypothesis to develop an analogical hypothesis that performs effectively in both domains \cite{liu2018few}. Transfer learning also has a high risk of generalization errors, which occur when certain classes of samples are more likely to appear in the target domain but rarely occur in the source domain \cite{liu2022few}. A model type that works to mitigate this is instance-based transfer learning, which focuses on selecting the most relevant examples from the source domain for training in the target domain. By effectively assigning weights to labeled instances in the source domain, this approach ensures that the instance distribution in both domains aligns closely, leading to a more reliable learning model with higher classification accuracy in the target domain \cite{liu2022few}. Another difficulty arises when there is insufficient overlap between the feature spaces of the source and target domains, making it essential to extract meaningful features that facilitate knowledge transfer \cite{liu2022few}. Feature-based transfer learning algorithms focus on finding the common feature representations between source and target domains, using them for knowledge transfer.

A third issue that arises is that in deep learning, there are few large-scale datasets that can train neural networks, which in turn results in poor classification results. To address this, fine-tuning transfer learning methods involve initially training a model on a large dataset, using the learned weights as the starting point for a new task with a smaller dataset, and then retraining the model with a reduced learning rate \cite{liu2022few}. In research in image classification in the medical field, this issue has been prevalent. For example, many models are trained on large datasets such as ImageNet before being fine-tuned on more specialized datasets to enhance their adaptability to medical imaging tasks \cite{cai2020few}. In the case of MR images, fine-tuning a pre-trained network with the data is often used as the quantity of data is low, and so overfitting must be addressed \cite{talo2019application}.

\subsubsection{Metric-Learning}\label{image_a3}
\begin{figure}[hbt]
\includegraphics[width=4.5in]{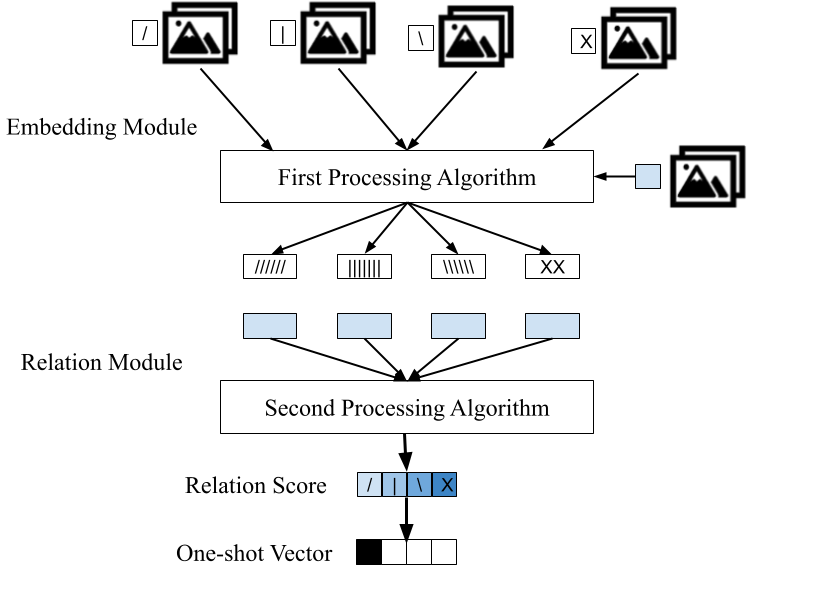}
\caption{Metric Learning Approach for Images}
\label{fig:MetricAudio}
\end{figure}

A third approach that can be taken with few-shot learning in the image-based domain is metric learning. Metric learning-based approaches focus on learning a set of projection functions so that, when images are represented in the embedding space, they can be easily classified using simple nearest neighbor or linear classifiers \cite{sung2018learning, matsumi2021few, lee2021metric, jiang2020multi, hao2019collect}. For metric learning to be successful, the samples of the same category (homogeneous) should be close together, while samples of different categories (heterogeneous) should be far apart \cite{li2020rs, li2023few}. This is accomplished by embedding data samples into a lower-dimensional latent space, reducing the distance between similar samples while increasing the separation between dissimilar ones \cite{prokop2023deep}. Past work in metric-based few-shot learning focused on using pre-defined distance metrics such as Euclidean or cosine distance for classification. However, some newer research has proposed that using a flexible function approximator to learn similarity can be beneficial, such as with a Relation Network. This new approach using a function approximator allows for a metric to be chosen based on the data, thus reducing the need to manually select a specific distance measure, such as Euclidean, cosine, or Mahalanobis \cite{sung2018learning}. Additionally, more recent metric-based few-shot learning techniques utilize neural networks to compare feature similarities between query and support samples \cite{li2020revisiting}. An overview of the metric-learning approach is shown in the diagram in {Figure~\ref{fig:MetricAudio}}. The samples are passed into the first processing algorithm, which processes them and passes them to the second processing algorithm with the baseline sample. The relation score is calculated and compared to the original sample to identify how the novel target is classified. 

\begin{equation*} 
\mathbb{D}_{\text{new}} = \left\{ (\hat{A}_i, \hat{B}_i) \mid \hat{A}_i \in \mathcal{A}_{\text{new}}, \hat{B}_i \in \mathcal{B}_{\text{new}} \right\}_{i=1}^{M_{\text{new}}} 
\tag{7}
\end{equation*}

The equation above represents the novel dataset utilized to collect and label new samples for the few-shot learning algorithm. In the equation $\hat{A}_i$ represents a new input sample, taken from the space $\mathcal{A}_{\text{new}}$, $\hat{B}_i$ represents the corresponding label which is taken from the space $\mathcal{B}_{\text{new}}$, $M_{\text{new}}$ represents the total number of new examples and the set notation $\left\{ (....)\right\}$ defines the collection of labeled pairs which are indexed from ${i=1}$ to $M_{\text{new}}$.

Each input sample has a corresponding label in the set \cite{zhang2024sample}. The data is sent through a processing algorithm such as ones for Prototypical networks, Siamese networks, and Matching networks in order to give the novel class samples a weighted metric to compare with the metric from the query set or image. 

There are multiple metric learning-based methods that can be applied. Typically, images are first used to create a feature space by embedding them into a model \cite{lu2024metric}. Different methods accomplish this in distinct ways, as the prototypical networks establish category prototypes within the feature space, while matching networks utilize attention mechanisms to construct efficient learning networks \cite{lu2024metric}. The second part is in distinguishing categories by distance metric, which the different methods process in different ways. The matching network applies an end-to-end trainable k-nearest neighbors algorithm on the learning embedding of a few labeled examples (support set) to classify unlabeled samples (query set) \cite{li2020revisiting}. The prototypical network builds a pre-class prototype representation. The relation network, which was previously mentioned, utilizes a shallow neural network to learn a nonlinear distance metric \cite{li2020revisiting}. 

Metric learning has taken on many forms over time, in different algorithms. Li et al. proposed a triplet-like metric learning approach named the Deep K-tuplet Network, an approach that extends the traditional triplet network by enabling joint comparisons with K negative samples within each mini-batch \cite{li2020revisiting}.  Another method proposed is a discriminative deep metric learning (DDML) method which was applied for face and kinship verification. Unlike many existing metric learning methods, the model utilizes a neural network to learn a series of hierarchical nonlinear transformations that project face images into discriminative feature subspaces, effectively addressing both nonlinearity and scalability challenges simultaneously  \cite{lu2017discriminative}. Tests conducted with this method indicated that the mean verification rates out performed previously proposed discriminative shallow metric learning methods. While the DSMML achieved an average accuracy of 0.92077 $\pm$ 0.004 in the unrestricted setting, DDMML achieved an accuracy of 0.9450 $\pm$ 0.0035. Another method, multi-scale metric-learning (MSML), extracts multi-scale features and captures relationships between samples at different scales for few-shot classification \cite{jiang2020multi}. When tested on mini ImageNet and tiered ImageNet, it was demonstrated that the MSML yielded superior performance, with an intra-class and inter-class relation loss function introduced to minimize distances between samples of the same class while enhancing the distinction between heterogeneous groups \cite{jiang2020multi}.  Another model that was proposed is the Coarse-to-Fine Metric-based Auxiliary (CFMA). This method was proposed to learn a distance metric for computing the similarity between sample pairs, even when the training and testing class sets are disjointed \cite{li2022coarse}. This method allows for the enhancement of the cross-domain transfer ability of the few-shot classification model, making it versatile. 

Metric learning has many benefits, but it also has some costs and challenges. Although it can adapt rapidly to new tasks, the process itself can be computationally expensive. Furthermore, metric-learning algorithms can often have high computational costs as a result of the frequent feature computations that take place and the nearest neighbor searches \cite{suh2019stochastic}. When the model is being tested, comparing the test data gathered against training data points can be expensive both computationally and in the resources required \cite{min2017exemplar}. 

\subsection{Time and Resource Considerations}\label{image_Costs}
As previously addressed, few-shot learning has many different models that can be selected for application in the image domain. While each of the methods has unique challenges and opportunities, they also have unique costs. Although meta-learning can be useful, its costs must first be evaluated. In some domains, data is limited but annotation of even limited data requires experts and results in high costs \cite{lu2023medoptnet}. Meta-learning can also be difficult to scale due to the high computational and memory costs, training instability, and lack of efficient distributed training support \cite{choe2023making, kaushik2022iterative}.

 Fine-tuning approaches can rely on large pre-trained feature extractors, and take many optimization steps when testing to learn tasks, resulting in a high computational cost for each new task \cite{bronskill2021memory}. When CNNs are used, there is also a risk of large processing time consumption to fine-tune without the use of a GPU \cite{shermin2018transfer}. It can also be very expensive to update a larger pre-trained model's parameter set \cite{sung2022lst} to fine-tune it. 

Metric-learning has training costs comparable to standard supervised learning on a base dataset such as ImageNet. The computational expenses depend on the complexity of the loss function applied, such as the K-tuplet loss which increases with the number of negative samples considered \cite{li2020revisiting}. Studies do suggest an optimal balance at K=5 to avoid any excessive costs. The method can adapt to new tasks efficiently, as it requires no further training --- only distance computations between query and support images in the learned space. The low adaptation cost makes metric-learning highly resource-efficient for inference, particularly in real-time image applications.

Few-shot fine-tuning transfer learning has a high initial pre-training cost but low adaptation cost, making it ideal for image tasks where pre-trained models can be reused across similar domains. Meta-learning offers flexibility for rapid adaptation, but its computational complexity during training can be a bottleneck. Metric-learning balances moderate training costs with the lowest adaptation overhead, and is ideal for cases where inference speed is important.

\subsection{Applications}\label{image_Applications}
\begin{itemize}
    \item  Few-shot learning has been applied in the image domain in medical imaging \cite{feng2021interactive, ding2023few, deo2024few, medela2019few, somani2024blend}. The medical use cases include the detection of diseases in the aforementioned images, which doctors or analysts may miss. Few-shot learning becomes relevant as diseases take many forms. Some diseases or scan results may be rare enough that there are no large databases of images to be processed by the machine learning model \cite{prabhu2019few, ouyang2022self}. As such, a few-shot learning model that accounts for smaller datasets and is trained to identify novel tasks using only a few labeled samples would serve well. 
    \item Wildlife and biodiversity monitoring and identification is another field where few-shot learning in the image domain can be considered beneficial. With its usage of smaller datasets and adaptability to small datasets, few-shot learning models can be used to identify rare species of plants, animals, insects, and more \cite{rodriguez2023fine, dong2024developing, miao2023few, nguyen2024art}. Camera traps are some of the more widely used tools to capture shots of wildlife because they do not disturb them, but these camera traps cannot take many images \cite{zhang2023few}. This makes images of wildlife limited, and using small datasets and few-shot learning a necessity.  
    \item Few-shot learning in the image domain can be applied in instances with facial recognition. While studies presently focus on facial expression recognition and other tasks, the research could be applied to facial recognition for security purposes \cite{shome2021fedaffect, yang2023two}. A user would input a few images of themselves into a small database to be used for verification and authentication. Few-shot learning would allow for more accurate verification when a user attempted to gain access to the device, as it would have adapted the smaller database to provide more accurate results than a larger database would have achieved. Previous works have noted that few-shot face recognition has been progressing toward replicating human visual intelligence, enabling efficient face recognition even after a single exposure \cite{holkar2022few}.
\end{itemize}

\section{Text or Natural Language Processing Domain}\label{Text}
The section below will present a detailed review of the challenges and opportunities of few-shot learning approaches in the text and natural language processing domain. 

\subsection{Introduction to Few Shot Learning Approaches and NLP}\label{NLP_Intro}
Few-shot learning is applied to many different tasks in the text and NLP domains. Although they are language-based, there are a vast number of different problems that they are involved in. For example, some research has indicated a belief that due to sparsity, sentences may be too short to contain sufficient information for labeling or distinguishing between different sentences in the distribution feature space \cite{yan2018few}, which is where few-shot learning could come into play. It is noted that to train good quality models it can take a large number of examples, which makes it a challenge to rapidly develop and deploy new models for real-world applications and needs \cite{muller2022active}. In the context of the text and NLP domain, the prompt-based method, embedding-based method, and data augmentation method will be discussed in the next sections. 

\subsection{Few Shot Learning Approaches for NLP}\label{NLP_Approach}
The paper will next cover approaches for few-shot learning in the NLP or text domain. The first approach will be Prompt-Based Learning in Section~\ref{NLP_a1}, followed by Embedding-Based learning in Section~\ref{NLP_a2}. The last approach will be Data Augmentation in Section~\ref{NLP_a3}. The applications on few-shot learning in the NLP/text domain will be covered following the approaches.

\subsubsection{Prompt-based Learning}\label{NLP_a1}
\begin{figure}[hbt]
\includegraphics[width=4.5in]{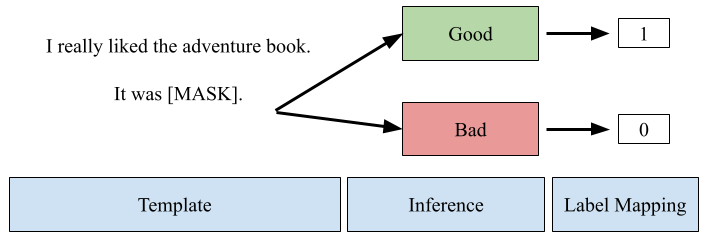}
\caption{Prompt-based Learning Approach for Text and NLP}
\label{fig:PromptBase}
\end{figure}

An issue with existing text classification methods is that methods which are based on external knowledge depend on large-scale training data to structure the model, leading to high costs for collecting suitable training samples and poor performance in few-shot learning \cite{wang2023knowledge}. Another issue with the current pre-trained models is that they require additional fine-tuning to solve few-shot text classification problems, yet with the small number of labeled training samples available in few-shot tasks, achieving a good fit is difficult, resulting in poor generalization performance \cite{liu2024knowledge}. Prompt-based few-shot learning is a natural language processing (NLP) technique that is gradient-free and uses only a few examples in the language model (LM) context \cite{madotto2021few, xu2022gps}. This approach relies solely on textual input provided for each sample in order to teach the LMs. This approach takes a language model trained to predict a word for a gap in a text and then transforms it into a modeling prompt using a template, querying a fine-tuned language model to fill it in, and then mapping it to a predicted class \cite{mayer2023prompt, thaminkaew2024prompt}. It has been noted that prompt-tuning has proved in recent works to be an effective few-shot learning approach to bridging the gap between pre-trained language models and downstream tasks \cite{zhang2022prompt, song2022investigating, le2023log}. Prompt-based learning has also been applied to solve the problem with keeping conversational models current with new conversational skills. This method is particularly advantageous as gradient-based fine-tuning is not required and a few examples in the LM context are instead use as the learning source \cite {madotto2021few}. A simplified diagram of the prompt-based template is shown in {Figure~\ref{fig:PromptBase}}. A template turns in the input into a text string with the mask (marked in the figure). The learning model fills the slot, and the words are mapped to the original labels.

\begin{equation*} 
\hat{z} = g \left( \arg\max_{z \in \mathcal{Z}} p(z \mid P(X); \phi) \right) 
= \arg\max_{g(z) \in \mathcal{Z}'} p(\text{[MASK]} = g(z) \mid P(X); \phi) 
\tag{8}
\end{equation*}
In the equation, $X$ represents the input text, $P(X)$ represents the prompt template which modifies the input text by inserting additional context tokens, including the $\text{[MASK]}$ token. Furthermore, $\mathcal{Z}$ is the original label space, and $g: \mathcal{Z} \rightarrow \mathcal{Z}'$ is the verbalizer which maps each label $z \in \mathcal{Z}$ to a subset $\mathcal{Z}'$. The section $p(z \mid P(X); \phi)$ is the probability assigned by the model to the label $z$ given the prompt-augmented input and the section $p(\text{[MASK]} = g(z) \mid P(X); \phi)$ is the probability that the verbalized token will be predicted at the $\text{[MASK]}$ position. 

The equation above demonstrates how the input is transformed into a prompt with a [MASK] token. A verbalizer function then maps each label from the original label space to a corresponding word, after which the model predicts the most probable filler for [MASK] through computation and selecting the label whose token has the highest probability \cite{yu2022unified}. 

Many models have been created using the prompt-based approach. The Prompt-Based Meta-Learning model was created to overcome the meta-learning problem of requiring too much data. The PBML model employees a two-level learning strategy, training base learner son label-specific vocabulary and a meta-learner on the overall prompt structure \cite{zhang2022prompt}. Throughout the research, the data indicated that using a prompting mechanism allows PBML to maintain strong performance even with substantially less meta-training data \cite{zhang2022prompt}. The Prompt-Learning for Short Text classification (PLST) model is one which intends to combat the issue of most prompt-learning methods, which often rely on simplistic strategies for incorporating knowledge, such as using only class names and a single approach for cloze-style predictions, leading to incomplete or biased results \cite{zhu2023prompt}. The model is comprised of three components to achieve this: concept retrieval, verbalizer construction, and text classification. With these components working together, the PLST model is intended to improve short text classification by considering the short text itself and the class name when generating label words, thus aiming to outperform existing methods \cite{zhu2023prompt}. When tested against five different datasets with 5-shot, PLST performed with an $F_1$-Score ranging from 0.2922 to 0.8259, and in all but one case achieving a higher $F_1$-Score than other models. Another proposed model which mitigates the limitations of traditional classification methods is the prompt-based Chinese text classification framework. The framework uses a template generation module and a demonstration filtering module, combining cosine similarity and mutual information into a new join correlation scoring function \cite{song2022investigating}. This joint scoring is noted to be key to the model's success. However, the experiment does note that finding suitable prompts is difficult and carefully designing the initialization prompts is critical to the model's success \cite{song2022investigating}. 

If multiple prompts are employed in the few-shot learning context, there can be higher training costs and longer interference times as a result of the assembly of predictions from various models \cite{nookala2023adversarial}. Previous research has also noted that prompt-based learning can be susceptible to the intrinsic bias which is present in pre-trained language models. This can result in sub-optimal performance in a few-shot learning setting \cite{he2024prompt}. 

\subsubsection{Embedding-based}\label{NLP_a2}
\begin{figure}[hbt]
\includegraphics[width=4.5in]{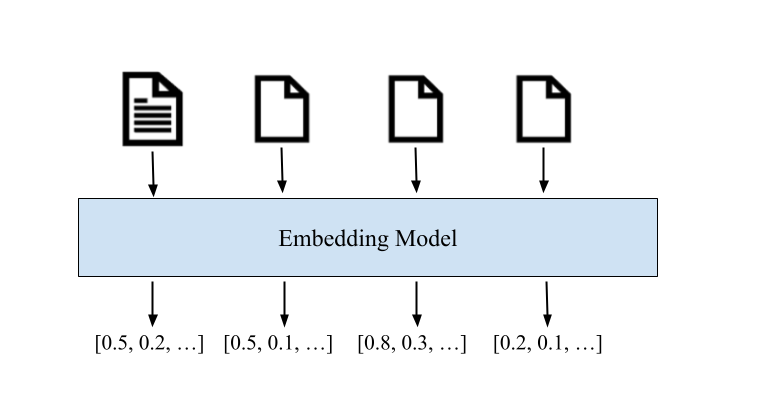}
\caption{Embedding-based Approach for Text and NLP}
\label{fig:EmbeddedText}
\end{figure}

Another approach to few-shot learning in the text and NLP domains is an embedding-based approach. The embedding-based approach is a type of transfer learning, making it able to work with few-shot learning to make it possible to learn from a few examples due to its learned representation \cite{ye2020few}. There are two main types of embeddings, word embeddings and sentence embeddings. It is important to know that word embeddings represent words as low-dimensional vectors of real numbers, capturing the semantic relationships between them \cite{bailey2018few}. Sentence embeddings are similar to word embeddings and represent sentences numerically. Tools such as the sentence encoding module of TensorFlow-hub can be used to generate high-dimensional sentence embeddings for NLP tasks such as determining semantic similarity and classifying text \cite{li2023siakey}. Word embedding is believed to be useful for a variety of NLP tasks, including named entity recognition, parsing, semantic relation extraction, and sentiment analysis \cite{zhu2023prompt, roy2019incorporating, ma2018concept, das2017named}. A simple flow of how the embedding-based model would function is shown in {Figure~\ref{fig:EmbeddedText}}. Each sentence passed in would be assigned values, and the closest value between the target and sample set would indicate successful classification.

\begin{equation*} 
\bar{h}^{mean} = \frac{1}{N} \sum_{j=1}^{N} e_j \tag{9}
\end{equation*}
In the equation, $\bar{h}^{mean}$ is the final mean-pooled sentence embedding, the $\frac{1}{N}$ averages the sum of the word embeddings with $N$ representing the total number of word embeddings, and $\sum_{j=1}^{N} e_j$ is intended to iterate over all the word embeddings in the sequence with $e_j$ as the embedding of the $j^{th}$ word in the sequence. The equation above is an example of the application of mean pooling in embedding-based few-shot learning. It computes the average of a sequence of word embeddings, which results in a fixed-size representation for a sequence. The mean pooling representation is then used as new examples are classified by comparing them to the fixed-size representation and making a decision based on their proximity \cite{pan2019few}.

One proposed model updates only the embedding vectors of a word by training on sentences that contain the word, along with negatively sampled sentences from prior network experiences \cite{lampinen2017one}. This method significantly decreases perplexity in text incorporating the new world while maintaining the model's understanding of other words, thereby preserving contextual integrity \cite{lampinen2017one}. Another model, which was proposed by Hu et al., was an adaptation for industrial alarm floods that combined word embedding and few-shot learning. The word embedding was used to convert alarm messages into word vectors that would be used in the model, and a long short-term memory method of FSL was adapted to identify the cause of the alarm floods \cite{hu2023root}. The results of the experiment indicated that the model was able to identify the root causes of the alarm floods with high accuracy, even with limited data. The skip-gram and Continuous Bag of Words (CBOW) models were implemented and trained on LSTM in this case, with the skip-gram model being adopted as the primary classifier, as it is more sensitive to words with lower frequencies. The results with a window size of 3 had a 0.903 accuracy while CBOW had an accuracy of 0.845.
Another proposed model, which takes on multiple languages for keyword identification, is the model presented by \cite{mazumder2021few}, which trains an embedding model on keyword classification using a multilingual, crowd-sourced speech dataset from Common Voice \cite{mazumder2021few}. The result of the work noted that incorporating surrounding audio context when training on keywords enhances the embedding model's ability to distinguish keywords despite coarticulation effects in speech \cite{mazumder2021few}.

However, flaws have been noted in the use of word-embeddings in first-generation pre-trained models, as while the models effectively capture semantic information, they often overlook the linguistic meaning of words and the contextual nuances embedded within them \cite{li2023siakey}. However, second-generation PLMs, such as BERT, GPT-2, and T5, incorporate contextual information, enabling them to grasp and interpret complex word concepts more effectively \cite{li2023siakey}. Another issue that may occur is that of out-of-vocabulary words. Most existing approaches operate under the assumption that each word appears frequently enough in the training corpus to derive a reliable representation from its surrounding context \cite{hu2019few}; however, this is not always true. The solution to this has arisen by approaching it as a few-shot regression, where an embedding prediction model, a hierarchical context encoder, and a meta-learning algorithm are adapted to resolve the issue of obscure words being difficult to assign a vector. 

\subsubsection{Data Augmentation}\label{NLP_a3}
\begin{figure}[hbt]
\includegraphics[width=4.5in]{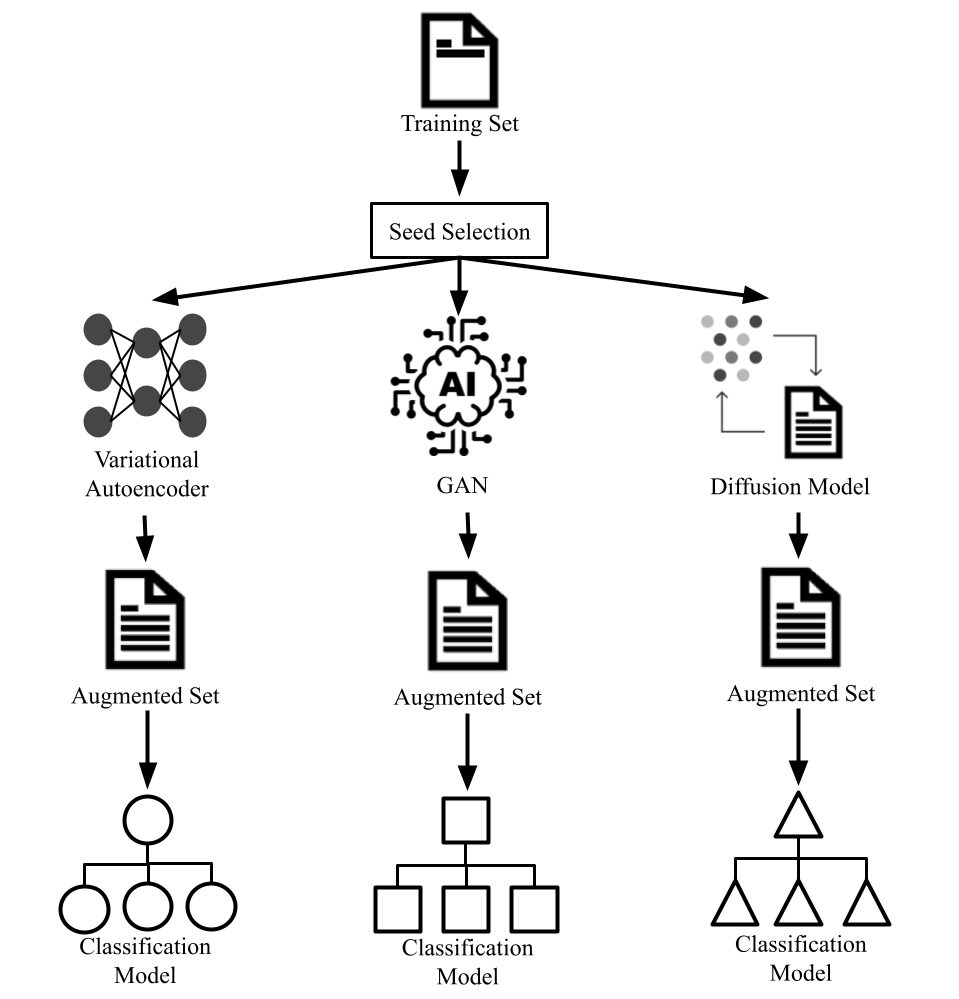}
\caption{Data Augmentation Approach for Text and NLP}
\label{fig:DataAug}
\end{figure}

Data augmentation is a method with which a training set is improved by generating new data to improve classifiers by creating the data artificially based on the given data \cite{zhou2021flipda, bayer2023data}. Importantly, it is usually model-agnostic and independent of any specific model architecture, making it highly versatile and applicable across various tasks \cite{dai2023auggpt}. In the case of text, the tasks that can be integrated into augmentation frameworks include replacement, insertion, deletion, and shuffling. Many works concentrate on image classification, and unlike images, text classification has more diverse expressions, which limits meta-learner performance and may necessitate a new model. Data augmentation is a proposed solution that generates additional samples to increase the number of training samples for novel categories while also mitigating the overfitting issues commonly associated with meta-learning-based few-shot learning approaches  \cite{sun2021meda, zhang2022cloze, abonizio2021toward, bencke2024data}. In past research, it has been noted that while data augmentation may provide only marginal improvements when ample training data is available, it becomes particularly advantageous in scenarios with limited data \cite{wei2021few}. There are many ways of introducing data augmentation into different models and tasks. Data augmentation in the feature space has been shown to enhance intent classification accuracy in few-shot settings beyond conventional transfer learning methods \cite{kumar2019closer}. Another method utilizes a cloze-style task, where a model predicts a hidden state vector as an augmented representation of an input sentence rather than reconstructing a complete sequence of tokens \cite{zhang2022cloze}. Further implementations of data augmentation are discussed next, and a flowchart diagram of how data augmentation works is shown in {Figure~\ref{fig:DataAug}}. The training set is passed into the seed selection and then passed to different generative AI models. From there, an augmented set is created using the training set, and it is sent through an appropriate classification model. 

\begin{equation}
L_{\text{gen}}(\Delta) = \frac{1}{2} \left( \alpha^T \alpha + \sum \left( \exp (\beta) - \beta - 1 \right) \right)
\tag{9}
\end{equation}
In the equation above, $L_{\text{gen}}(\Delta)$ represents the loss function for the generative model, which ensures that the learned latent space is structured for effective sample generation. The $(\Delta)$ represents the KL divergence, which measures the difference between the standard and learned latent distribution. The $\frac{1}{2}$ is the scaling factor to simplify gradient calculations when optimizing the loss, and in $\alpha^T \alpha$, $\alpha$ represents the mean of the latent space distribution while the dot product overall computes the squared magnitude of the mean vector. The section $\sum \left( \exp (\beta) - \beta - 1 \right)$ is the variance regularization term. $\beta$ represents the logarithm of the variance of the latent space distribution, $\exp (\beta)$ ensures the variance remains positive, and subtracting $\beta - 1$ ensures the loss remains formulated correctly for the KL divergence. 
Equation 9 above shows a basic equation demonstrating the application of the Kullback-Leibler divergence loss applied to a generative model \cite{liu2024load}. It is used to regulate latent space distribution which ensures the encoded feature distribution aligns with the prior normal distribution. This is essential to ensuring the sample generation is both stable and diverse. 

The FlipDA model is proposed to achieve effectiveness and robustness for hard few-shot tasks such as textual entailment and word sense disambiguation. The model first generates augmented data by substituting words using a pre-trained T5 model, followed by a classifier that selects data with altered labels \cite{zhou2021flipda}. The label-flipped data was key to the augmentation procedures as it helps reduce the model's reliance on spurious correlations \cite{zhou2021flipda}. The researchers noted that preliminary experiments indicated the label-flipped data improved the generalization of pre-trained models when compared to augmented data, which preserved the original labels. Overall, when the labels were flipped with data-augmentation applied as well, the performance on the tasks tested improved by up to 10 points. For example, with the CB-F1 task, the accuracy at baseline was 0.7216 $\pm$ 0.0027, the accuracy with no label flipping and only data augmentation was 0.7707 $\pm$ 0.0491, while the accuracy with DA and label flipping was 0.8814 $\pm$ 0.0393. Another proposed model is one that combines data augmentation and randomly selected training sentences for the purpose of machine translation. For each reference sentence pair provided that contained the word of interest, sentences from the training corpus with similar contexts around the word of interest would be identified \cite{arthaud2021few}. The second step would then be to insert the word of interest and it's translation into the retrieved sentences. This methodology resulted in a very high BLEU score and the highest accuracy improvements \cite{arthaud2021few}. Another metric to note is that the model reported better accuracy scores with only 1 to 5 examples than a system that was trained with 313 parallel examples on average. Yet another model which was proposed, the AUGNLG model, which is a novel data augmentation approach that generates Meaning Representation (MR)-to-text data automatically from open-domain text \cite{xu2021augnlg}. This is accomplished by combining a self-trained neural retrieval model with a few-shot learning-based Natural Language Understanding (NLU) model. In this case, the data augmentation process consists of retrieving utterances that match relevant keywords, filtering out domain-irrelevant instances, and generating synthetic MR annotations to enhance training efficiency \cite{xu2021augnlg}. 
In some cases, data augmentation may be generated with rules created by hand. However, these rules are based on expert domain knowledge and can require a high cost of labor. Augmentation rules may apply to specific problems, which makes it more difficult to apply the rules to other tasks \cite{cho2022improving}. 

\subsection{Time and Resource Considerations}\label{NLP_Costs}
Few-shot embedding-based learning pre-trains models on large corpora, which is a costly process that can take many days on GPUs. Adaptation in this case is efficient and makes use of fine-tuning embeddings or lightweight classifiers with minimal resources, making it effective when pre-trained embeddings are available. Few-shot prompt-based learning uses large language models that are pre-trained on large datasets with high computational expenses \cite{brown2020language}. Adaptation in this case requires no extra training and relies on in-context learning within prompts \cite{gao2020making}. However, inference is resource-heavy, scaling with model size and prompt complexity. Few-shot data augmentation generates synthetic text, and the costs vary from low for simple methods to higher for generative models like T5 \cite{raffel2020exploring}. Here, adaptation involves training on augmented data, which is a moderate expense compared to pre-training, and balances flexibility and resource use. Overall, embedding-based learning is best suited for pre-trained scenarios, prompt-based learning is suitable for rapid deployment, and data augmentation is best when there is limited data and more is needed. 

\subsection{Applications}\label{NLP_Applications}
\begin{itemize}
    \item Any application involving crossing language barriers would benefit from the application of few-shot learning \cite{mozafari2022cross, zaharia2020cross, qiu2024cross, he2020multi, shafikuzzaman2024empirical}. With Cross-Lingual Text Classification, there is a lack of data, making classification difficult without the use of manual labor \cite{feng2024prompt}. Vocabulary differences between languages make cross-lingual knowledge transfer challenging; however, pre-trained language models and few-shot-based models could make a difference in these classification attempts and knowledge transfer attempts, as models based on few-shot learning are intended to adapt to the small amount of data in order to be successful. 
    \item Few-shot learning can also be applied to text classification. One such classification is sentiment analysis, as sentiment in a sentence is often indicated within the context, and varies from sentence to sentence \cite{zhou2024soft, taher2021adversarial}. This results in a smaller sample size to be digested. Furthermore, the classifying or labeling of many words would be a daunting and time-consuming task \cite{zhao2024multi}, which could be resolved with few-shot learning models applying techniques such as data augmentation. Topic categorization is another classification task that few-shot learning could be applied to. The model could be applied to pull words from the article and assign the article a category based on the score of different words in the title, without having to analyze the entire article. 
    \item Few-shot learning in the text or NLP domain can also be applied to intent recognition \cite{wu2024triple, kumar2021protoda, luoyiching2024relation}. With only a few examples of user intent, the model would be able to recognize a broad range of similar commands. This would allow for greater functionality without having to have a large volume of labeled data for the model to sort through. 
\end{itemize}

\section{Miscellaneous: Multi-Modal Scheme}\label{multi}
    The section below will review the challenges and opportunities in the multi-modal domains, such as audio and image domains, or text and image domains. 
    
\subsection{Introduction to Multi-Modal Scheme}
Multi-modal scenarios in few-shot learning involve training models to perform tasks with a limited number of samples. This is similar to other domains in few-shot learning; however, in the multi-modal domain, multiple modalities such as text, image, and audio are leveraged to train the model to perform the task. It has already been noted that collecting large amounts of labeled data in real-world applications can be difficult and expensive. For this, few-shot learning has emerged as a solution. A new challenge arising is that current few-shot methods struggle to process the growing variety of multi-modal data, such as multi-spectral images or multimedia content, which differ significantly from the widely studied RGB data used in most research \cite{yang2024cross}. Multi-modal data, which is collected from different sources and modalities, is a proposed solution for the problem. In the section below, one approach to the multi-modal scheme will be further discussed, along with some existing and proposed applications.

\subsection{Multi-Modal Few-Shot Learning}
Multi-modal few-shot learning was introduced, in part, because few-shot setups have ambiguity, depending on the task. Lin et al. argued that when a learner is presented with an image, it cannot know what it should be identifying in the image \cite{lin2023multimodality}. Multi-modal learning, however, will allow for the different modalities to work together to improve the visual classifier in such cases. Notably, multi-modal learning is designed to address the challenge of transferring knowledge across different modalities \cite{cheng2023causal} and may, when necessary, combine meta-learning and transfer learning in a single framework to meet this goal. In a number of current works, the focus has been on the combination of text and image classifiers as computer vision systems develop \cite{wang2023few, lin2023multimodality, zhang2024cross}. In some cases, multi-modality involves not only crossing domains but also crossing methodologies. For example, a model that proposes the combination of text classification and image classification for object detection can also use prompt-based few-shot learning and meta-learning few-shot learning \cite{han2022multi}. Another common multi-modal domain intersection is that of the video and image domains. In these scenarios, multi-modal modeling can lead to various downstream tasks, such as video-text retrieval, image-text retrieval, and visual question answering for both images and videos \cite{lu2023uniadapter}.

\begin{equation}
\begin{aligned}
w_c &= \sum_j \beta_j y_j \varphi (x_j) = \sum_n W_n c, \\
\text{where} \quad W_n c &= \sum_{j: \varphi_j=n} \beta_j c \varphi (x_j)
\end{aligned}
\tag{10}
\end{equation}

Above, $w_c$ represents the final weight vector for class $c$, which is obtained by combining the training samples from all of the modalities. The $\sum_j \beta_j y_j \varphi (x_j)$ represents the classifier's weight, which is formed by a weighted sum of the training features, $x_j$, where $\beta_j$ is a learned coefficient. The section $\sum_n W_n c$ represents the total weight which is the sum of the modality-specific weights, where $n$ is the set of all modalities. The section $W_n c = \sum_{j: \varphi_j=n} \beta_j c \varphi (x_j)$ computes the modality-specific weight by taking the sum of all the training samples belonging to the modality $n$.
With multi-modal learning, classifiers are produced that are groups of modality-specific classifiers. These classifiers support the application of the Representer Theorem to represent classifiers as "linear combinations of their training samples" \cite{lin2023multimodality}. The equation above represents the weights for the class $c$, which are a weighted combination of all $j$ training features across all modalities. This joint optimization improves adaptation by considering all modality-specific weights. 

\begin{figure}[hbt]
\includegraphics[width=4in]{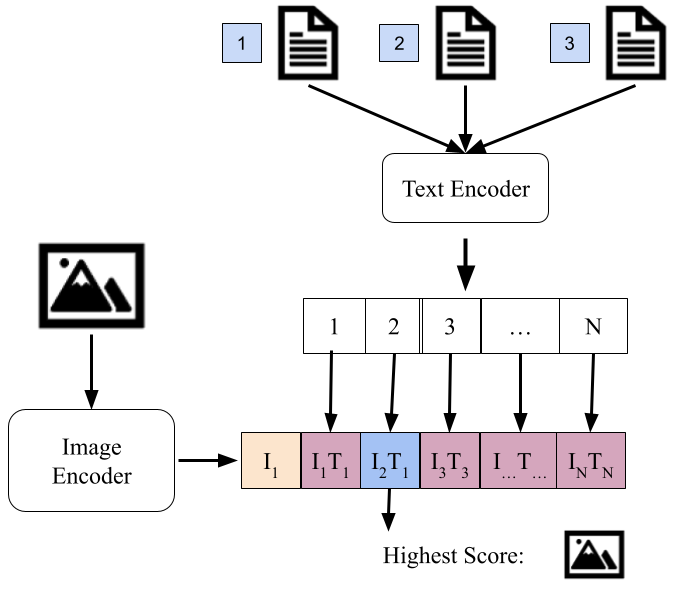}
\caption{Multi-modal Few-Shot Learning with Text and Image Domains}
\label{fig:MultiModal}
\end{figure}

In one proposed model, multi-modal transfer learning was approached to assist with more accurate image classification. In this case, images and text were combined in the model. Chen et al. noted that this approach incorporates multi-modal contextual information to leverage heterogeneous data more effectively, enhancing similarity measurement between few-shot samples and the target domain while mitigating biases from individual samples \cite{chen2021cross}. A diagram of how the image and text domains are combined is shown in {Figure~\ref{fig:MultiModal}}. Both are passed through their own encoders, which normalize their scores for comparison. The highest merged score is then the 'answer'. Different text samples and an image are encoded and scored. The scores are combined using an algorithm, and the highest score combination between the text and image indicates that the corresponding text and image are related. Another model took the approach of cross-domain contrastive learning (XDCL), which was intended to learn hyperspectral image representations. These representations cross domains as the hyperspectral data captures the spectral and spatial domains, each of which represents the same object in different ways. Since hyperspectral data captures both spectral and spatial domains, this method predicts whether two representations from different domains correspond to the same sample, effectively capturing shared features between them. The multi-modal task for XDCL was used to predict which representations of different domains match the same sample \cite{guan2022cross}, thus capturing what is shared between the domains. Yet another proposed model is the multi-modal contrastive learning module (CCLM), which is designed to improve the learning of feature representation by using semantic information at the class level \cite{wang2024cross}. In the case of this model, action recognition in videos is the specific task, and the model is intended to address the multi-modality of the video and image domains. The paper noted that the proposed multi-modal contrastive learning network (CCLN) outperformed other state-of-the-art few-shot action recognition methods. When compared to state-of-the-art few-shot action recognition methods, CCLN performed with an accuracy of 0.875 in the 5-shot testing, while the other methods performed with accuracy in the range of 0.753 - 0.874, with most falling below 0.874.

\subsection{Time and Resource Considerations}\label{multi_Costs}
Multi-modal few-shot learning boosts classification by integrating different modalities such as text and images, but can be computationally expensive \cite{xing2019adaptive}. In some cases, separate encoders for each modality and joint optimization of weights demands significant memory and processing power, especially with large datasets \cite{chen2021cross}. In the case of combining meta-learning and transfer-learning, costs are increased further with iterative training and high-dimensional data processing, requiring robust GPU/TPU support. The need to preprocess heterogeneous data for score normalization also adds more overhead. 

\subsection{Applications}

\begin{itemize}
    \item Multi-modal few-shot learning has been applied to action and gesture recognition. In terms of action recognition, the video and image domains intersect to allow for action detection in videos such as playing a cello. This has also been applied to skeleton action detection, where skeleton sequences and RGB videos were considered two different views of the same action and two different domains \cite{lu2024cross}. In the case of gesture recognition, the image and acoustic domains have been applied together in order to better detect what gestures are being made \cite{han2022multi}.
    \item Another application of multi-modal few-shot learning is hyperspectral image classification. Hyperspectral data is limited, but with multi-modal models, other modalities such as RGB images, textual data, and pre-trained embeddings can be applied to improve visual classifiers. In some cases, the combination of image and text features is claimed to produce an improved visual understanding, allowing for stronger generalization across different modalities \cite{li2024cross}. This can be applied in situations such as tree species classification, where linguistic features of class names are used to help refine the decision boundaries of visual classifiers \cite{hu2024cross}.
    \item Multi-modal few-shot learning can also be applied to the medical field. One example with which this could be achieved would be with the classification of rare diseases by transferring knowledge from text annotations to images. Report generation from medical images using only a few annotated examples could be another application of multi-modal few-shot learning in the medical field. 
\end{itemize}

\section{Discussion}

In the age of increasing use of big data, the general perception is that models trained on large datasets outperform models trained with smaller datasets, i.e., the power of a model greatly depends on the volume of data used to train the model. While this assumption holds most often due to the enhanced generalizability of a machine learning model developed with a large dataset with variations of patterns compared to its smaller counterparts. But finding a large usable dataset is not always easy due to several factors, including time and the nature of the data.  

To resolve such challenges, a few-shot learning-based model development could be an alternative. 
Few-shot learning (FSL) can be applied across multiple domains, such as image, audio, and text/NLP domains. It can also be applied in cases where multiple domains can be combined. To select which methodology suits a situation best, it is important to understand the benefits and functionality of each, including potential problems, which are thoroughly presented in this manuscript. 
For instance, while in the text and NLP, few-shot-based approaches, such as embedding-based FSL and prompt-based FSL, could ease the data and resource demand made by popular large language models (LLMs), a multi-modal FSL can combine domains such as text and image, audio and image, text and audio, and more to develop predictive models with a smaller dataset.


While few-shot learning approaches bring the benefit of achieving a reasonable performance through developing models with a smaller dataset, they have some constraints, including time and resource requirements. For instance, while the model-agnostic meta-learning (MAML) approach in the audio domain can achieve a high performance, it has a higher computational cost. On the other hand, the dynamic few-shot learning approach in the audio domain may require more resources. It is pivotal for future researchers to consider all these factors across the domain-varying approaches while developing their few-shot-based models, depending on their situation and need. Thereby, this manuscript presents comprehensive details to guide future directions in this area of research. 

\ignore{-------------------
Many different methods can result in higher or lower computational and resource costs depending on circumstances. Each method has its advantages and disadvantages. Some methods, such as MAML, have higher computational costs, while others may have higher resource costs. Which should be selected varies by both domain and goal.

With respect to the current age of LLMs, where more data is considered better for accurate results, more research could be conducted to test the effectiveness of few-shot methodologies. Depending on the application needed, different few-shot methodologies in the same domain could be tested and compared to determine which methodology applies best in which scenarios. Deploying NLP/text-based few-shot learning models to be used in addition to larger ones, such as GPT, would help users gain experience with an LLM trained on limited data. This would challenge he myths that for an LLM to be effective you need an extremely high quantity of samples. 
-------------------}

\bibliographystyle{IEEEbib}
\bibliography{reference_Andrea,reference_self_full}
\end{document}